# Deep Learning Approach for Matrix Completion Using Manifold Learning

Saeid Mehrdad, Mohammad Hossein Kahaei

*Abstract*—Matrix completion has received vast amount of attention and research due to its wide applications in various study fields. Existing methods of matrix completion consider only nonlinear (or linear) relations among entries in a data matrix and ignore linear (or nonlinear) relationships latent. This paper introduces a new latent variables model for data matrix which is a combination of linear and nonlinear models and designs a novel deep-neural-network-based matrix completion algorithm to address both linear and nonlinear relations among entries of data matrix. The proposed method consists of two branches. The first branch learns the latent representations of columns and reconstructs the columns of the partially observed matrix through a series of hidden neural network layers. The second branch does the same for the rows. In addition, based on multi-task learning principles, we enforce these two branches work together and introduce a new regularization technique to reduce over-fitting. More specifically, the missing entries of data are recovered as a main task and manifold learning is performed as an auxiliary task. The auxiliary task constrains the weights of the network so it can be considered as a regularizer, improving the main task and reducing over-fitting. Experimental results obtained on the synthetic data and several real-world data verify the effectiveness of the proposed method compared with state-of-the-art matrix completion methods.

## I. Introduction

WHEN the missing entries of a low-rank partially observed matrix are recovered using its observed entries, it refers to Matrix Completion (MC) [1-4]. MC arises in various study fields including image and video denoising [5, 6], image inpainting [7, 8], collaborative filtering [9, 10], etc. The MC methods try to recover the missing entries of a low-rank partially observed matrix, $Y \in \mathbb{R}^{m \times n}$, using its observed entries $Y_{i,j}(i,j) \in \Omega$, where $\Omega$ is a matrix which indicates the positions of observed entries of $Y$. Under the assumption that complete matrix is a low-rank matrix, the nuclear-norm and Matrix Factorization (MF) based approaches are two popular categories for MC. Nuclear-norm based approaches are based on rank minimization. As the rank minimization problem is non-convex and generally NP-hard [11], nuclear-norm minimization is used in this category as a convex approximation of rank minimization [12, 13]. In [7, 14-16] a few extension of nuclear-norm are employed to make a better approximation of rank function. For example, the Truncated Nuclear-Norm (TNN) as a better approximation to the rank of matrix than the nuclear-norm is considered in [7]. The Hybrid Truncated Norm regularization (HTN) method which is a combination of truncated nuclear norm and truncated Frobenius norm is addressed in [15] for MC. The TNN model based on weighted residual error is discussed in [16]. A combination of MF based and nuclear-norm based approaches for MC which drawn the data from multiple subspaces in the presence of high coherence is presented in [2]. In MF based approaches, the complete matrix is approximated by the multiplication of column and row latent variables [17]. Therefore, in the last decade, MF approaches are widely used in MC tasks due to its efficiency in latent features learning. [9, 18, 19]. The main idea of MF is to factorize the observed matrix into two (or more) low-rank latent variables as follows:

$$Y = UV, \qquad (1)$$

where $U \in \mathbb{R}^{m \times r}$, $V \in \mathbb{R}^{r \times n}$, and, $r$ is the rank of $Y$ that should be estimated or given in advance. The missing entries of $Y$ can be recovered through finding optimal values for $U$ and $V$ to approximate the observed matrix [17, 20-22]. Much research efforts has been done to improve the efficiency of MF approaches in MC tasks. For example a Confidence-aware Matrix Factorization (CMF) framework for recommender systems is given in [19]. The accuracy of rating is optimized and the prediction confidence is measured simultaneously in CMF. In [18], an Extended-Tag-Induced Matrix Factorization technique for recommender systems is addressed. In this method, the correlation between tags derived by co-occurrence of tags are used to enhance the performance of recommender systems. An evolutionary approach called Evolutionary Matrix Factorization (EMF) is presented in [9]. The EMF produces matrix factorizations automatically in order to improve the performance of recommender systems.

Non-convexity of objective function and sensitivity to the rank estimation of completed matrix are the major drawbacks of MF based methods. In nuclear-norm based methods, the objective function is convex and estimated or given rank is not required. However, the computational complexities in nuclear-norm minimization is higher than MF methods. The reason is that the major computation of the MF methods is the multiplication of two latent factors in each iteration but in nuclear-norm minimization computing a partial Singular Value Decomposition (SVD) is required in each iteration. Computing the SVD causes a high computational cost particularly in large-scale MC tasks such as recommender systems. Therefore in these cases MF based methods are often applied [23]. In addition, low-rankness is based on linear latent variable models [24]; and thus, MF based and nuclear-norm based methods provide linear reconstruction of observed data. Consequently, their performance decrease when the data are from nonlinear latent variable models [25-28] as follows:

$$Y = f(V), \qquad (2)$$

where $f(.)$ is a nonlinear function. To recover the missing entries of incomplete matrix $Y$ given by (2), the nonlinear function $f(.)$ and the latent factors $V$ are needed to be approximate.

Recently, deep neural networks have been successful in several study fields due to their efficiency in learning latent features and powerful data representations [29-36]. They can also approximate $f(.)$ via nonlinear activation functions. Therefore, in recent years, these methods are widely used in nonlinear MC [27, 28, 30, 37]. The first autoEncoder based method for nonlinear MC called AutoEncoder based Collaborative Filtering (AECF) is addressed in [38]. AECF takes the partially observed matrix as the input, replaces the missing entries by pre-defined constants, projects it onto a latent space through an encoder layer, and, reconstructs it via a decoder layer. The biases introduced by the pre-defined constants affect the performance of AECF. A representation learning framework called Recommendation via Dual-autoencoder (ReDa) is presented for recommender systems in [37]. In ReDa, by applying autoencoders, the latent representations of users and items are learned simultaneously. In this method, the deviations of training data by the learnt representations of users and items is also minimized to find better latent representations. Another autoencoder based method, in which the partially observed data is incorporated to present a deep neural network framework is given in [27]. In [28], a Deep Matrix Factorization (DMF) approach is addressed for nonlinear MC. In DMF, the inputs are unknown latent variables and the output is the partially observed data. By optimizing the inputs and parameters of the deep neural network, reconstruction errors for the observed entries are minimized. Next, missing entries are obtained by propagating the latent variables to the output layer. A patch-based nonlinear matrix completion method is discussed in [39]. In this method, by employing a convolutional neural network, the predictive relationship between a matrix entry and its surrounding entries is learned through an end-to-end trainable model that leads to a nonlinear MC solution.

Indeed, the aforementioned methods try to find better and more general latent factors to make more accurate reconstructions of the partially observed data matrix. In these models, the latent variables are assumed to locate in nonlinear subspaces and nonlinear latent variables are learned; hence, these models are restricted to handling matrices with nonlinear relations among all its entries and linear relations are ignored. But real datasets do not have completely nonlinear relations among their entries. Thus, these methods do not provide general latent variables; although, they make better reconstructions of the partially observed data than their linear counterparts. On the other hand, deep neural network based methods bear the risk of over-fitting, particularly, when the observed matrices are highly sparse such as recommender systems [30]. To handle this, $\ell_2$ regularization such as weight decay and dropout are applied in [29, 40, 41]. Side information is also used in [40-43] to increase the available information data for training. The performance of $\ell_2$ regularization methods reduce under high sparsity settings [29, 30] and side information is not available in all datasets (e.g., in image inpainting and classification).

In this paper, we propose a new latent variables model for the partially observed data matrix in which both linear and nonlinear relations among the entries of the data are considered. The proposed method is a combination of linear and nonlinear models introducing in (1) and (2). To address both linear and nonlinear relations among entries of data matrices, we design a novel deep neural network based method in this work. In addition, by considering multi-task learning, we also introduce a new regularizer technique to overcome over-fitting,

The proposed method is a two-branch deep neural network. The first branch is a column-base deep neural network which is trained to learn the latent variables of columns of the observed matrix. It takes the low-dimensional unknown latent variables as inputs and approximates both nonlinear and linear relations among entries of the data matrix via a deep neural network. The nonlinear relations are approximate through a series of hidden neural network layers with nonlinear activation functions as in [28].We also enforce the network weights and the inputs to approximate the linear relations in parallel with nonlinear part via linear activation function. Then the combination of linear and nonlinear effects is considered as the reconstruction of observed matrix. In contrast to existing methods, by introducing a combination of linear and nonlinear models for the data, this model does not restrict the latent variables to locate in nonlinear (or linear) subspaces. Therefore through this parallel structure, this framework learns both linearity and nonlinearity from the data. The network is trained and the network parameters are optimized to minimize the reconstruction errors for the observed entries that are the outputs of the network. iRprop+ is employed to solve the optimization problem. The second branch is a row-based deep neural network that performs a similar process on the rows of the data matrix and learns the latent variables of rows.

In this model, the multiplication of the weights of column-based branch (row-based branch) are considered as rows (columns) latent variables. To handle the over-fitting problem, we design a hybrid model by combining the column-based and row-based branches and minimize the deviations of observed entries by the multiplications of the weights of the two branches as a manifold learning objective. The proposed method is based on the multi-task learning rules in which the missing entries of data are recovered as a main task and manifold learning is performed as an auxiliary task. The auxiliary task constrained the weights of the network; thus, it can be considered as a regularizer, leading to mitigate the over-fitting and improving the main task. The efficiency of the manifold learning in over-fitting reduction is shown in [30, 44]. Actually, instead of exploiting side information that is not available in all datasets, we extensively exploit the observed data, approximating more general latent variables and also decreasing the over-fitting. The proposed method is compared with deep neural network based methods, MF based methods, and, nuclear-norm based methods for MC. The experimental results on different datasets demonstrate that our proposed method has superiority over all of the compared methods in reconstruction accuracy. The regularization quality and the influence of the introduced regularizer in over-fitting reduction is also evaluated.

The remainder content of this paper is organized as follows: Section II presents the details of our proposed method. In section III several experiments are conducted in the tasks of synthetic matrix completion, image inpainting, and, collaborative filtering. Section IV concludes our paper.

## II. PROPOSED APPROACH

To consider both linear and nonlinear relations among observed matrix entries, a two-branch deep neural network framework is designed in this paper. The first branch makes reconstructions of the columns of the observed matrix and in a similar way, the second branch makes reconstructions of the rows of the observed matrix. Furthermore, we enforce these two branches work together using multi-task learning structure and introduce a regularization term to mitigate over-fitting. For brevity, we only explain the column-based network in detail; the row-based network can be similarly formulated.

### A. Model structure

A new latent variables model is proposed in this paper. The proposed model is a combination of models in (1) and (2) as follows:

$$Y = f(V) + UV + \epsilon, \tag{3}$$

where $Y \in \mathbb{R}^{m \times n}$ is the observed matrix, $U \in \mathbb{R}^{m \times r}$, $V \in \mathbb{R}^{r \times n}$ are the latent representations of rows and columns respectively, $\epsilon$ is an additive noise, and, $r = Rank(Y) < min(m, n)$. Therefore $Y$ is redundant ($r < m$) and the missing entries of $Y$ in model (3) can be recovered. To recover the missing entries of an incomplete matrix in model (3), we need to approximate $f(.)$, $U$, and, $V$. To achieve this, we minimize the difference between observed entries of $Y_{i,j}$ and $[f(V) - UV]_{i,j}$ where $(i, j) \in \Omega$ and $\Omega$ indicates the locations of observed entries of $Y$. Therefore, the latent representations of rows and columns ($U$ and $V$) and the nonlinear function $f(.)$ can be learned via solving the following minimization problem:

$$\min_{f,U,V} \frac{1}{2n} \| I \odot (Y - f(V) - UV) \|_F^2, \tag{4}$$

where the Hadamard matrix product is shown as $(A \odot B)_{i,j} = A_{i,j} B_{i,j}$, $\| \cdot \|_F$ denotes the Frobenius norm, and, $I \in \mathbb{R}^{m \times n}$ is an indicator matrix where $I_{i,j}$ is 1 if $Y_{i,j} \neq 0$ and 0, otherwise. We design a column-based deep neural network structure to solve (4). The deep neural network can generate better representations of arbitrary nonlinear relations among the input entries and outperform the shallow structures [45, 46]. Fig. 1 illustrates the architecture of the column-based branch. It takes the unknown column latent factors ($V$) as input and make an approximation of nonlinear term in (3) through hidden layers with nonlinear activation functions as in [28]. In parallel with the nonlinear part, we also enforce the network weights to make an approximation of the linear term in (3) by applying linear activation function. Therefore, the nonlinear and linear parts are approximated as follows:

$$f(v) = \sigma^{(L+1)}(W_c^{(L+1)} \sigma^{(L)}(W_c^{(L)} \ldots \sigma^{(1)}(W_c^{(1)} v + b_c^{(1)}) \ldots) + b_c^{(L)}) + b_c^{(L+1)}), \tag{5}$$

$$Uv = W_c^{(L+1)}(W_c^{(L)} \ldots (W_c^{(1)} v + b_c^{(1)}) \ldots) + b_c^{(L)}) + b_c^{(L+1)}. \tag{6}$$

where $L$ is the number of hidden layers, $v$ is a column of $V$, $W_c^l$ denotes the weight matrix, $b_c^l$ denotes the bias vector of the $l$-th layer, $l = 1, 2 \ldots, L + 1$, and, $\sigma^{(l)}(.)$ is the nonlinear activation function of the $l$-th layer. Sigmoid function, hyperbolic tangent function, and, Rectified Linear Unit (ReLU) are widely-used activation functions. According to (6), $U$ is approximated by the product of the column-based network weights as follows:

$$U = \prod_{l=1}^{L+1} W_c^l. \tag{7}$$

By substituting (5) and (6) into (4), the objective function to learn the latent representations of columns ($V$) and the network parameters can be rewritten as follows:

$$\mathcal{L}_{column} = \frac{1}{2n} \| I \odot (Y - \sigma^{(L+1)}(W_c^{(L+1)} \sigma^{(L)}(W_c^{(L)} \ldots \sigma^{(1)}(W_c^{(1)} V + B_c^{(1)}) \ldots) + B_c^{(L)}) + B_c^{(L+1)}) - W_c^{(L+1)}(W_c^{(L)} \ldots (W_c^{(1)} V + B_c^{(1)}) \ldots) + B_c^{(L)}) + B_c^{(L+1)}) \|_F^2, \tag{8}$$

where $n$ columns of bias vector $b_c^l$ form $B^{(l)}$. $f(.)$ and $V$ can be computed by solving (8). Therefore, we make reconstructions of the columns of the observed matrix $Y$. In contrast to existing methods of matrix completion tasks that assume the observed data as a nonlinear (or linear) transformations of lower-dimensional latent representations, both linear and nonlinear relations are learned from the data in our method and we do not restrict the latent factors to just handling nonlinear (or linear) relations among the entries of the data. The missing entries can be given by

$$Y_{i,j} = [f(V) - UV]_{i,j}, (i,j) \in \overline{\Omega}, \tag{9}$$

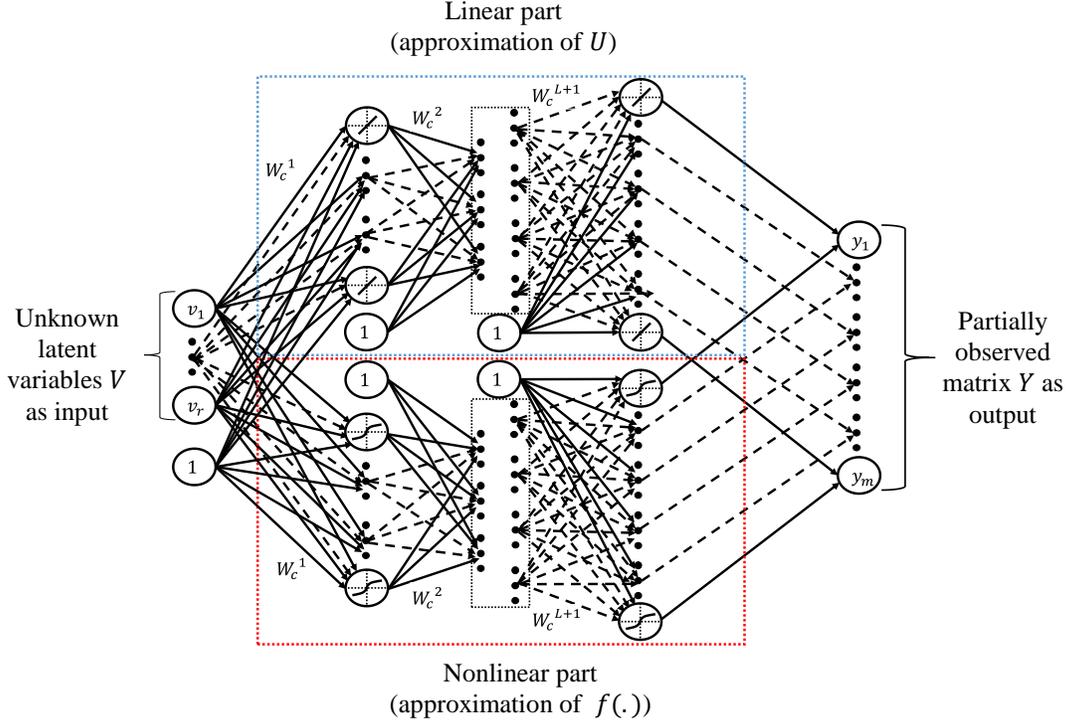

Fig. 1. Diagram of the column branch of the proposed network.

where the positions of missing entries are indicated by $\bar{\Omega}$.

Similarly, the row-based branch takes the unknown row latent factors ($U$) as input and generates reconstructions of the rows of observed matrix as output. In row-based network, the latent representation of columns ($V$) can be approximated by the product of the networks weights as follows:

$$V^\top = \prod_{l=1}^{L+1} W_r^l. \tag{10}$$

The objective function to learn row latent factors $U$, nonlinear function, and, network parameters in row-based network is defined as follows:

$$\mathcal{L}_{row} = \frac{1}{2m} \| I^\top \odot (Y^\top - \sigma^{(L+1)}(W_r^{(L+1)} \sigma^{(L)}(W_r^{(L)} \ldots \\ \sigma^{(1)}(W_r^{(1)} U^\top + B_r^{(1)}) \ldots) + B_r^{(L)}) + B_r^{(L+1)}) - \\ W_r^{(L+1)}(W_r^{(L)} \ldots (W_r^{(1)} U^\top + B_r^{(1)}) \ldots) + B_r^{(L)}) \\ + B_r^{(L+1)}) \|_F^2, \tag{11}$$

where $W_r^l$ and $B_r^l$ denote the weights matrix and bias matrix of the $l$-th layer respectively, and, $(.)^\top$ indicates the transposition of the matrix. $U$ can be computed by solving (11). We now have two independent branches that work separately. To mitigate over-fitting, a regularization term is introduced that enforce these two branches work together. Multi-task learning rules are used to impose proposed regularization term.

Multi-task learning is to use information comes from other related tasks to enable our model to perform better on our main task [44, 47]. It is common in multi-task learning to share representations between related tasks. By learning in each task, learning in other tasks is improved. Single-task learning models suffer the risk of over-fitting to this task. But multi-task learning models introduce an inductive bias by the auxiliary task as a regularizer to mitigate the over-fitting risk. Learning more tasks implicitly increases the available training data and enforces the model to learn a representation that includes all of the tasks; therefore, the risk of over-fitting in the main task is reduced [44]. In our model, the main task generates column and row reconstructions of the observed matrix. We also constrain the approximations of latent representations of column and rows obtained from (7) and (10), respectively, to operate as an auxiliary task by realizing the matrix factorization model (1). Following this idea, we minimize the deviations of observed matrix by the learned latent representations in (7) and (10) as follows:

$$\mathcal{L}_{reg} = \frac{1}{2n} \| I \odot (Y - (\prod_{l=1}^{L+1} W_c^l)(\prod_{l=1}^{L+1} W_r^l)^\top \|_F^2. \tag{12}$$

By applying (12) to train our model we enforce the two branches work together. Thus this model reconstructs the observed matrix as the main task and performs an auxiliary task simultaneously by employing the manifold learning objective in (12). The auxiliary task imposes constrain on the network weights so it can be considered as a regularizer that improves the main task and mitigate over-fitting. Fig. 2 illustrates the structure of the proposed framework. The Frobenius norm of

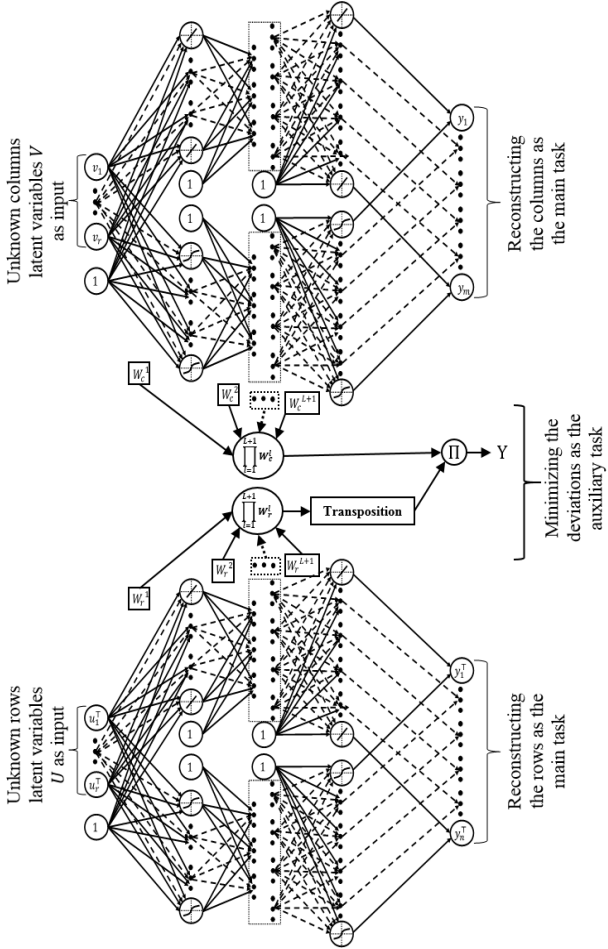

Fig. 2. Diagram of the proposed method.

the weights and inputs of the two branches given by:

$$reg(V, U, W_c, W_r) = \frac{1}{2n}\|V\|_F^2 + \frac{1}{2m}\|U\|_F^2 + \frac{1}{2}\sum_{l=1}^{L+1}\|W_c^l\|_F^2 + \frac{1}{2}\sum_{l=1}^{L+1}\|W_r^l\|_F^2 \quad (13)$$

are also used as regularization to drive the weights and inputs to decay which can be complementary to the proposed regularizer in (12).

### B. Objective function and optimization

In contrast to single-task learning methods in which one activation function is optimized, in multi-task learning the objective function consists of several components that are related to the main and auxiliary tasks. The objective function of proposed method is formulated as follows:

$$\mathcal{L} = \alpha \cdot \mathcal{L}_{column} + \beta \cdot \mathcal{L}_{row} + \gamma \cdot \mathcal{L}_{reg} + \lambda \cdot reg(V, U, W_c, W_r), \quad (14)$$

where $\alpha, \beta, \gamma$, and, $\lambda$ are appropriate hyperparameters that control trade-off between the main task loss terms and the regularization terms. The objective function in (14) is nonconvex and the inputs $(U, V)$ of the network are unknown; therefore, a nonlinear and batch-wise optimization technique is required to solve it. Hence we choose improved resilient back-propagation (iRprop+) to solve the objective function because iRprop+ is a nonlinear and batch-wise approach and its efficiency in deep neural network optimization has been proved in [48]. The gradient of the objective function is needed in iRprop+. We can compute $\partial \mathcal{L}_{column}/\partial V$, $\partial \mathcal{L}_{column}/\partial W_c$, $\partial \mathcal{L}_{row}/\partial U$, $\partial \mathcal{L}_{row}/\partial W_r$, $\partial \mathcal{L}_{reg}/\partial W_r$, and, $\partial \mathcal{L}_{reg}/\partial W_c$ by back-propagation algorithm. Thus the gradient of (14) can be readily computed (in each layer of the proposed network, bias vector can be considered as a column of the output of that layer; therefore, its gradient can be obtained through computing the gradient of the other parameters). We initialize the inputs ($V$ and $U$) and network parameters according to [28]. Therefore the inputs are set as zeros and network parameters are initialized as in [49]. It is worth noting that the data and the activation function outputs should be in the same range. The data can be easily transform into the appropriate interval and returned to its original interval after optimization to evaluate the reconstruction error.

## III. EXPERIMENTS

In this section, several experiments have been conducted to evaluate the performance of the proposed matrix completion scheme from two aspects: 1. Regularization quality and over-fitting reduction, 2. Accuracy in reconstructing the missing entries. The proposed method is implemented in synthetic matrix completion, image inpainting, and, collaborative filtering tasks and its performance is compared with the five state-of-the-art matrix completion methods. The compared methods are selected from MF based methods, nuclear-norm based methods, and, deep neural network based methods for MC. The first compared method is a MF based method solved by LMaFit [17]. The second one is AECF method [38] which is an auotoencoder based methods and solved by nonlinear conjugate gradient method. LRFD [2] which is a combination of MF based and nuclear-norm based methods is the third method. The fourth method is DMF method [28] solved by iRprop+ and BFGS. DMF is a deep neural network based method. The last one is a nuclear-norm based method called HTN [15] . The parameters of all these methods are adjusted as suggested by the authors or set carefully to provide the best performance. In proposed method, ReLU and sigmoid function are used as the activation function for the nonlinear part of the network. To successfully recover the missing entries, the input size $r$ (as the estimation of the data matrix rank) should be smaller than the number of observed entries in each vector of the data matrix. The hyperparameters $\alpha, \beta, \gamma$, and, $\lambda$ are selected from {0.01, 0.05, 0.1, 0.5, 1}.

The peak signal-to-noise ratio (PSNR) [15] is employed to evaluate the performance of proposed method for image inpainting, defined as:

TABLE I
THE COMPARISON OF DIFFERENT METHODS ON SYNTHETIC DATA

| Mask | Metric | LMaFit | AECF | LRFD | DMF | HTN | Proposed |
|---|---|---|---|---|---|---|---|
| 30% | PSNR | 17.9324 | 22.2136 | 18.2298 | 29.0314 | 19.4357 | **30.8654** |
|  | SSIM | 0.9029 | 0.9594 | 0.9402 | 0.9869 | 0.9522 | **0.9956** |
| 50% | PSNR | 17.0231 | 20.1489 | 17.2154 | 26.1489 | 18.7874 | **27.8231** |
|  | SSIM | 0.8886 | 0.9466 | 0.9337 | 0.9748 | 0.9456 | **0.9923** |
| 70% | PSNR | 16.4462 | 17.9841 | 16.8964 | 20.5412 | 17.3261 | **23.8934** |
|  | SSIM | 0.8779 | 09375 | 0.9201 | 0.9622 | 0.9489 | **0.9845** |

TABLE II
RESULTS (PSNR) OF THE DIFFERENT METHODS ON 9 RGB IMAGES WITH 30% RANDOMLY MASKED PIXEL. BOLDFACE NUMBERS DENOTE THE BEST PSNR

| Image | Size | LMaFit | AECF | LRFD | DMF | HTN | Proposed |
|---|---|---|---|---|---|---|---|
| Image-1 | 300×300 | 28.6214 | 28.4982 | 28.9812 | 29.7249 | 29.4814 | **30.2224** |
| Image-2 | 300×300 | 27.4987 | 27.5896 | 28.0019 | 28.0446 | 28.1953 | **28.3436** |
| Image-3 | 300×300 | 22.2753 | 23.1174 | 23.5154 | 23.8869 | 23.9105 | **23.9569** |
| Image-4 | 300×300 | 38.3134 | 38.9165 | 39.1298 | 39.5123 | 39.6312 | **40.0634** |
| Image-5 | 300×300 | 23.7259 | 23.8569 | 24.0173 | 24.9864 | 24.5758 | **25.3054** |
| Image-6 | 300×400 | 24.9857 | 24.9745 | 25.1174 | 25.7967 | 25.6492 | **26.1442** |
| Image-7 | 300×300 | 23.3914 | 23.6561 | 23.9859 | 24.8832 | 24.6841 | **25.0234** |
| Image-8 | 210×350 | 31.6721 | 31.7254 | 31.8983 | 32.2123 | **32.3673** | 32.3477 |
| Image-9 | 512×512 | 18.2258 | 18.5416 | 18.8347 | 19.1647 | 19.1484 | **19.5683** |

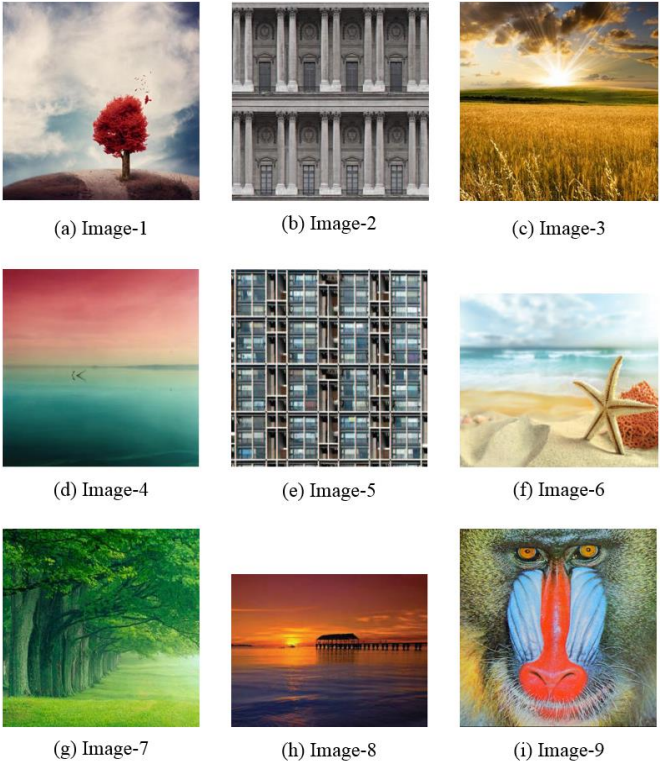

Fig. 3. RGB images for inpainting

TABLE III
COMPARISON OF DIFFERENT METHODS ON 6 IMAGES WITH RANDOM MASK, TEXT MASK, AND, BLOCK MASK

| Image | Mask | Metric | LMaFit | AECF | LRFD | DMF | HTN | Proposed |
|---|---|---|---|---|---|---|---|---|
| Image-1 | 30% | PSNR | 28.6214 | 28.4982 | 28.9812 | 29.6249 | 29.4814 | **30.2224** |
|  |  | SSIM | 0.8275 | 0.8269 | 0.8399 | 0.8629 | 0.8463 | **0.8722** |
|  | 40% | PSNR | 28.0039 | 27.9642 | 28.4469 | 29.3363 | 29.2281 | **29.8847** |
|  |  | SSIM | 0.786 | 0.7946 | 0.8074 | 0.8488 | 0.8131 | **0.8578** |
|  | 50% | PSNR | 27.5134 | 27.4329 | 28.0114 | 28.7921 | 28.6161 | **29.0411** |
|  |  | SSIM | 0.7301 | 0.7221 | 0.7398 | 0.8009 | 0.7588 | **0.8159** |
| Image-2 | 30% | PSNR | 27.4987 | 27.5896 | 28.0019 | 28.0446 | 28.1953 | **28.3436** |
|  |  | SSIM | 0.9306 | 0.9364 | 0.9499 | 0.9501 | 0.958 | **0.9671** |
|  | 40% | PSNR | 26.7435 | 26.7736 | 27.3164 | 27.3264 | 27.6341 | **27.7214** |
|  |  | SSIM | 0.8917 | 0.9084 | 0.9294 | 0.9291 | 0.9373 | **0.948** |
|  | 50% | PSNR | 25.9864 | 26.0324 | 26.5463 | 26.6429 | **26.8386** | 26.7984 |
|  |  | SSIM | 0.8664 | 0.8691 | 0.9021 | 0.901 | **0.9094** | 0.9089 |
| Image-3 | Text | PSNR | 24.3647 | 25.5686 | 25.9236 | 26.5236 | 26.5131 | **27.0162** |
|  |  | SSIM | 0.9377 | 0.9476 | 0.9509 | 0.966 | 0.9612 | **0.971** |
| Image-4 | Text | PSNR | 42.2086 | 42.5436 | 42.9802 | 43.1812 | 43.2281 | **43.4523** |
|  |  | SSIM | 0.9556 | 0.969 | 0.9732 | 0.9864 | 0.9887 | **0.9921** |
| Image-5 | Block | PSNR | 17.4632 | 17.7954 | 17.9495 | 18.9821 | 18.9014 | **19.3436** |
|  |  | SSIM | 0.9901 | 0.9916 | 0.9923 | 0.9957 | 0.9941 | **0.9977** |
| Image-6 | Block | PSNR | 16.0212 | 16.2293 | 16.3723 | 17.4129 | 17.3861 | **17.7963** |
|  |  | SSIM | 0.9722 | 0.9744 | 0.9759 | 0.9774 | 0.9771 | **0.9794** |

$$PSNR = 10\log_{10}\frac{mn(\max(Y))^2}{\|\overline{Y}-Y\|_F^2}, \quad (15)$$

where $Y$ and $\overline{Y}$ are the observed matrix and the recovered matrix, respectively, and, $\max(Y)$ is the maximum entry in $Y$. We also evaluate the quality of reconstruction by the structure similarity (SSIM) [15], defined as:

$$SSIM = \frac{(2\mu_{\overline{Y}}\mu_Y+C_1)(2\sigma_{\overline{Y}Y}+C_2)}{(\mu_{\overline{Y}}^2+\mu_Y^2+C_1)(\sigma_{\overline{Y}}^2+\sigma_Y^2+C_2)}, \quad (16)$$

where $\mu_Y$, $\mu_{\overline{Y}}$ are the mean values of $Y$ and $\overline{Y}$, and, $\sigma_Y$, $\sigma_{\overline{Y}}$ indicate the variances, respectively; $\sigma_{\overline{Y}Y}$ indicates the covariance between $Y$ and $\overline{Y}$. $C_1$ and $C_2$ are two constants that prevent the dominator from becoming zero. Larger values of PSNR and SSIM indicate more accurate recovery.

### A. Synthetic matrix completion

To evaluate the performance of the proposed method, we generate a synthetic dataset for matrix completion with a $m \times n$ matrix of rank $r$ as follows:

$$Y = g(1.2(0.5g^2(AB) - g(AB) - 1)) + AB, \quad (17)$$

where $A \in \mathbb{R}^{m \times r}$ and $B \in \mathbb{R}^{r \times d}$ have independent and identically distributed (i.i.d.) Gussian entries and $g(x) = 1.7159\tanh(\frac{2}{3}x)$ is element-wise activation function as in [27]. We set $m = 100, n = 200$, and, $r = 10$. Two hidden layers are employed in this set of experiments. The number of nodes in the input layer, hidden layers, and, output layer are set as [10 20 40 $m$] and [10 25 50 $n$] in column and row branches, respectively. The hyperparameters $\alpha, \beta, \gamma,$ and, $\lambda$ are adjusted empirically and tentatively as 1, 1, 0.01, and, 0.01, respectively. We randomly remove 30%, 50%, and, 70% of the entries. The average results of 50 repeated tests are reported in Table I. The results show that the proposed method outperforms other methods (especially linear methods) significantly.

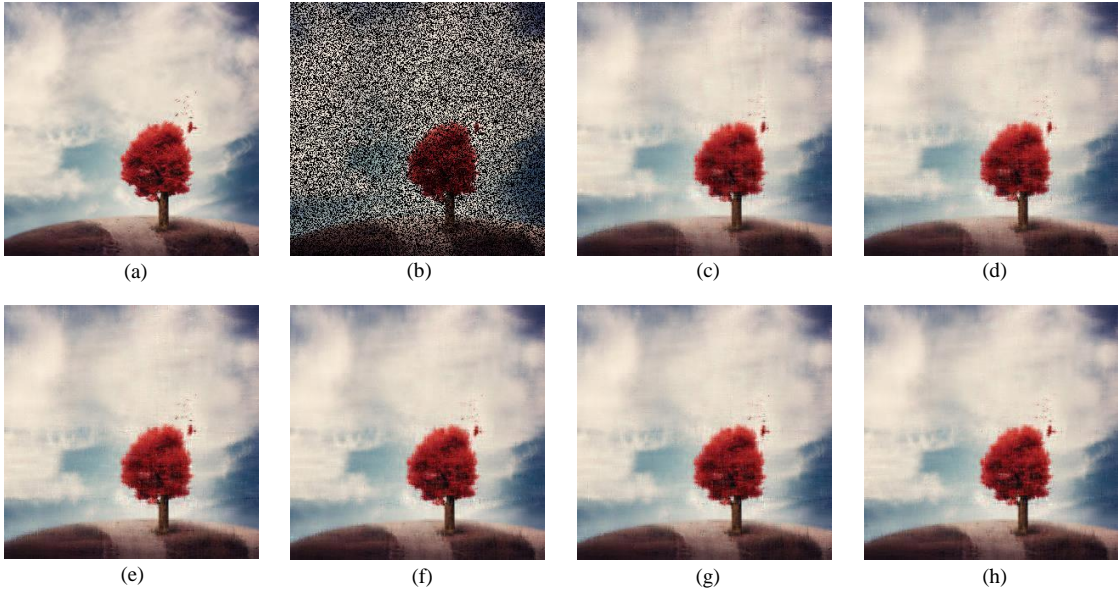

Fig. 4. Inpainting results for Image-1 with 50% random pixel mask. (a) Original image; (b) Masked image; (c) LMaFit; (d) AECF; (e) LRFD; (f) DMF; (g) HTN; (h) Proposed.

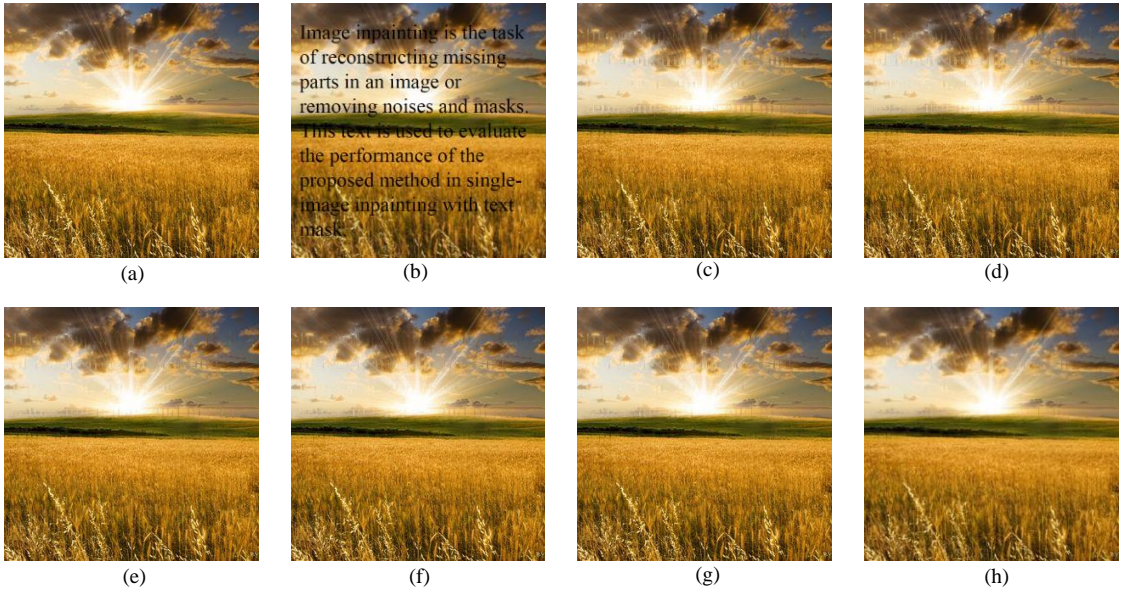

Fig. 5. Inpainting results for Image-3 with text mask. (a) Original image; (b) Masked image; (c) LMaFit; (d) AECF; (e) LRFD; (f) DMF; (g) HTN; (h) Proposed.

### B. Applications for single-image inpainting

Image inpainting is the task of reconstructing missing parts in an image or removing noises and masks [7]. Images can be considered as low rank matrices [15]; hence, matrix completion methods are used to reconstruct missing entries in images. We use a set of widely-used RGB images in image inpainting task, shown in Fig. 3, to measure the performance of proposed method. In the proposed method, one hidden layer is employed for each branch. The number of nodes in the input layer, hidden layer, and, output layer are set as $[r\ 200\ m]$ and $[r\ 400\ 3n]$ in column and row branches, respectively (The number of nodes in output layer of the row branch is $3n$ because the RGB images are unfolded to pixel matrices). The hyperparameters $\alpha, \beta, \gamma$, and, $\lambda$ are set as 1, 1, 0.01, and, 0.5, respectively, and, $r \in [15,100]$. For all images, a random mask which randomly removes 30% of the pixels is applied. The images sizes and the average PSNRs for 10 tests are summarized in Table II. It can be seen that in most cases the proposed method outperforms all the other methods. To further evaluate the efficiency of the proposed method, three types of masks are used. Two images are arbitrarily selected from the images set in Fig. 3 for each mask. For example, random pixel masks (with 30%, 40%, and, 50% randomly removed pixels) is applied for image-1 and image-2, text mask is applied for images-3 and image-4, and, block mask is employed for image-5 and image-6. For each case, we repeat the test 10 times and the average PSNRs and SSIMs are reported in Table III. The results show that the performances of the proposed method are the best in most cases. Moreover, Fig. 4, Fig. 5, and, Fig. 6 show some of the original images, masked images, and, reconstructed images given by different methods. As can be seen, the proposed method makes the best reconstructions of the original images.

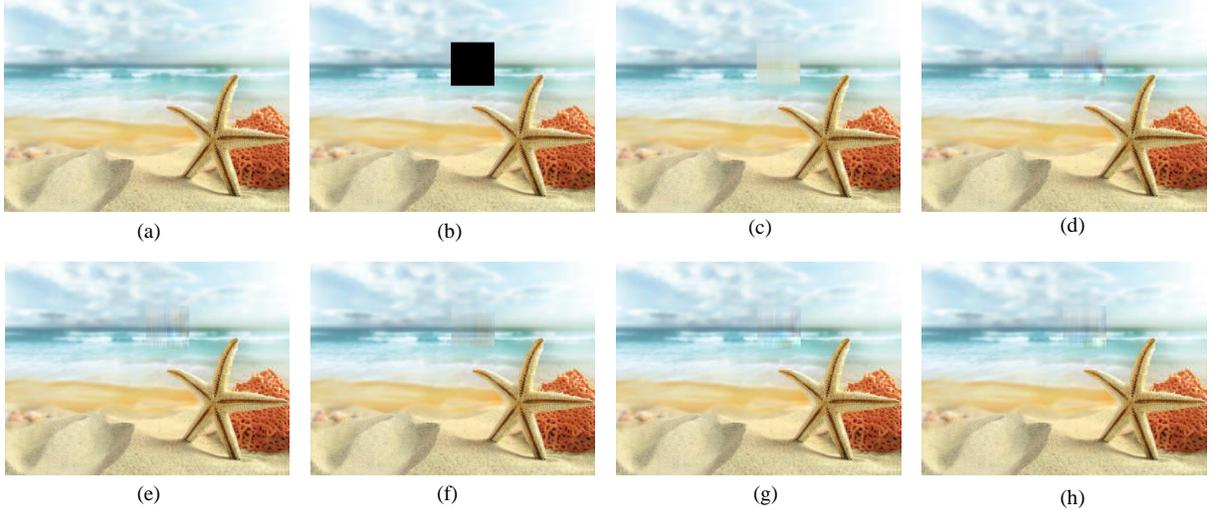

Fig. 6. Inpainting results for Image-6 with 60× 60 block mask. (a) Original image; (b) Masked image; (c) LMaFit; (d) AECF; (e) LRFD; (f) DMF; (g) HTN; (h) Proposed

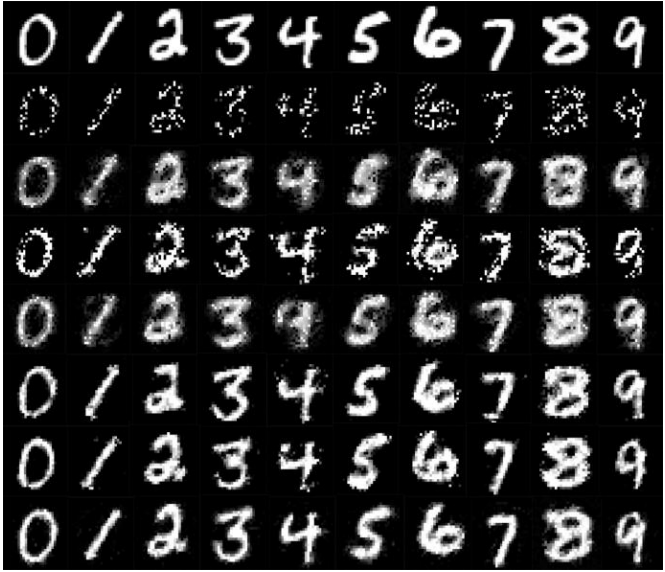

Fig. 7. Inpainting results for MNIST images with 70 % random pixel mask: row-1: original, row-2: masked, row-3: LMaFit, row-4: AECF, row-5: LRFD row-6: DMF, row-7: HTN, row-8: Proposed.

TABLE IV
THE COMPARISON OF DIFFERENT METHODS ON MNIST IMAGES WITH RANDOM PIXEL MASKS

| Mask | LMaFit | AECF | LRFD | DMF | HTN | Proposed |
|---|---|---|---|---|---|---|
| 30% | 15.0463 | 16.9418 | 16.5163 | 21.9467 | 18.1996 | **23.5957** |
| 50% | 14.4332 | 16.0211 | 15.8857 | 19.0123 | 17.1354 | **20.3749** |
| 70% | 13.9121 | 15.1116 | 14.9626 | 16.5124 | 16.0123 | **17.1136** |

### C. Applications for group-image inpainting

In order to evaluate the reconstruction performance of our method in group-image inpainting, The MNIST dataset of handwritten digits (28× 28) [50] is utilized. We form a 784× 1000 matrix by selecting 100 images for each digit randomly. Each image is a column of the matrix. Random pixel masks which randomly remove 30%, 50%, and, 70% of the pixels are applied in this study. One hidden layer is employed for each branch. The number of nodes in the input layer, hidden layer, and, output layer are set as [10 100 $m$] and [10 100 $n$] in column and row branches, respectively. The hyperparameters $\alpha$, $\beta$, $\gamma$, and, $\lambda$ are set as 1, 1, 1, and, 0.5, respectively. The visual results for 70% random pixel mask is shown in Fig. 7. As it can be observed, the proposed method provides more accurate reconstructions for digits and they have the least difference with the original digits compared to other methods. The average PSNRs for 10 tests are reported in Table IV. We can see that in all cases the PSNR values of the proposed method are larger, demonstrating the efficiency of the proposed method.

### D. Applications for collaborative filtering

In this study, we apply the proposed method to two well-known datasets for collaborative filtering. The MovieLens 100k[1] is the first one containing 100,000 ratings (1 to 5) for 1682 movies from 943 users. The second dataset is the Jester-joke dataset-1[2] collected from 24983 users who have rated 36 or more jokes of 100 jokes. The range of ratings is [-10, 10]. The first dataset forms a 943× 1682 rating matrix in which rows refer to users and columns refer to movies. We randomly select the ratings of 10000 users in Jester-joke data and form a $10000 \times 100$ matrix. For both datasets, we randomly sample 70% and 50% of the ratings for training respectively, and use the rest for evaluations. The performance of the proposed method for collaborative filtering is evaluated by widely-used normalized mean absolute error (NMAE) [27], defined as:

$$NMAE = \frac{1}{(Y_{max}-Y_{min})|\bar{\Omega}|} \Sigma_{(i,j) \in \bar{\Omega}} |\bar{Y}_{ij} - Y_{ij}|, \quad (18)$$

where $Y_{ij}$ and $\bar{Y}_{ij}$ are the true and recovered ratings of movie $j$ by user $i$, respectively, $|\bar{\Omega}|$ is the number of the missing entries, and, $Y_{max}$, $Y_{min}$ are the upper and lower bounds of ratings, respectively. In this set of experiments, one hidden layer is applied. The number of nodes in the input layer, hidden layer, and, output layer are set as [10 100 $m$] and [10 200 $n$] ([5 100 $m$] and [5 10 $n$]) in column and row branches for the MovieLens 100k (the Jester-joke), respectively. For Jester-joke the hyperparameters $\alpha, \beta,$ and, $\lambda$ are set as in Sec. III –A; but for MovieLens, $\gamma$ is set as 1 because the sparsity in the MowieLens dataset is much greater than the Jester-joke (about 72% of ratings are available in the Jester-joke dataset-1 and only 6.3% are in the MovieLens 100k). Therefore the regularization plays a more important role in the MovieLens 100k (see also Sec. III –E). The average NMAEs for 10 tests are summarized in Table V. The results illustrate that the proposed method predicts the missing ratings effectively for the all cases and outperforms the other methods.

---

[1] http://grouplens.org/datasets/movielens/100k/

[2] http://eigentaste.berkeley.edu/dataset/

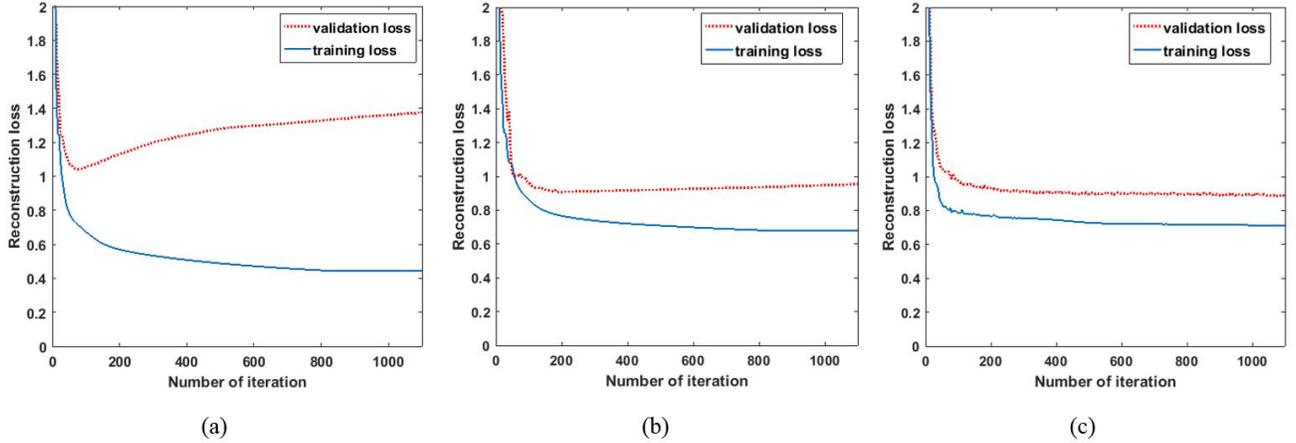

Fig. 4. Training and validation reconstruction losses of the column branch of the proposed method. (a) $\gamma = 0$ and $\lambda = 0$; (b) $\gamma = 0$ and $\lambda = 0.5$; (c) $\gamma = 1$ and $\lambda = 0.5$.

TABLE V
RESULTS (NMAE (%)) OF THE DIFFERENT METHODS ON THE JESTER-JOKE DATASET-1 AND MOVIELENS 100K

| Data | Mask | LMaFit | AECF | LRFD | DMF | HTN | Proposed |
|---|---|---|---|---|---|---|---|
| Jester-joke | 30% | 15.94 | 16.13 | 15.72 | 15.68 | 15.71 | **15.61** |
|  | 50% | 16.32 | 16.41 | 16.11 | 16.03 | 16.08 | **15.98** |
| MovieLens 100k | 30% | 18.97 | 19.02 | 18.64 | 18.25 | 18.61 | **18.02** |
|  | 50% | 20.17 | 19.29 | 18.96 | 18.73 | 18.92 | **18.51** |

*E. Regularization quality*

In this section, the influence of the auxiliary task (as a regularizer) on the reconstruction accuracy and over-fitting is investigated. The proposed method is implemented on the MovieLens 100k and Jester-joke dataset-1. We use 75% of the ratings for training the model, 20% for evaluation, and, 5% for validation. In the experiment, we give the set of $\gamma = \{0.01, 0.05, 0.1, 0.5, 1\}$ and keep the other three parameters fixed. The reconstruction errors for different values of $\gamma$ are reported in Table VI. As we can see, the lowest error for the MovieLens 100k and Jester-joke occurs when $\gamma = 1$ and $\gamma = 0.01$, respectively. Large value of $\gamma$ shows the importance of the regularization. As we claimed previously, the regularization effect increases when the number of training entries is small and the data matrix is sparse. This happens when we implement our model on MovieLens 100k.

TABLE VI
RESULTS (NMAE (%)) OF THE PROPOSED METHOD ON THE JESTER-JOKE DATASET-1 AND MOVIELENS 100K WITH DIFFERENT VALUES OF $\gamma$

|  | $\gamma = 0.01$ | $\gamma = 0.05$ | $\gamma = 0.1$ | $\gamma = 0.5$ | $\gamma = 1$ |
|---|---|---|---|---|---|
| Jester-joke | **15.43** | 15.44 | 15.47 | 15.54 | 15.57 |
| MovieLens 100k | 17.98 | 17.95 | 17.91 | 17.86 | **17.84** |

TABLE VII
RESULTS (NMAE (%)) OF THE PROPOSED METHOD ON THE MOVIELENS 100K WITH AND WITHOUT PROPOSED REGULARIZER

|  | $\gamma = 0$ | $\gamma = 1$ |
|---|---|---|
| Column branch | 17.97 | **17.84** |
| Row branch | 18.16 | **17.92** |

Now we evaluate the influence of the auxiliary task in over-fitting reduction. First both regularizer in (14) are removed by setting $\gamma$ and $\lambda$ as zero. In this case column and row branches of the proposed model work separately. The training and validation reconstruction losses on the MovieLens 100k are shown in Fig. 8. As can be seen in Fig. 8 (a), the model over-fitting is high in this case. Now we apply the regularizer in (13). We can see in Fig. 10 (b) that the generalization error; which is a gap between the training and the validation loss, becomes lower when the regularizer is employed. At the end we employ the auxiliary task as a regularizer by setting $\gamma = 1$. Fig. 8 (c) shows the generalization error decrease further when we have both regularizers and the best results are obtained in this case. Therefore different regularization approaches can work together as a complement. We also compare the outputs of the column and row branches with and without the auxiliary task, for $\gamma = 1$ and $\gamma = 0$. The results are reported in Table VII. We can find that the auxiliary task improves the main task and the reconstruction errors of the proposed method with the auxiliary task ($\gamma = 1$) is lower than the separate branches ($\gamma = 0$) in both column and row reconstructions of the observed matrix. It worth mentioning that column branch makes more accurate reconstructions because in this experiment we have more ratings per item than user.

## IV. CONCLUSION

In this paper, we introduce a new latent variables model for data matrix which is a combination of linear and nonlinear models and propose a regularized deep-neural-network-based method for to handle both linear and nonlinear relations among data matrix entries. In addition, we introduce a regularization term to mitigate over-fitting based on multi-task learning principles. Experimental results demonstrate that the proposed method outperforms the state-of-the-art matrix completion methods to recover the missing entries. We also evaluate the influence of the introduced regularizer on over-fitting reduction. The results verify the effectiveness of the introduced regularization technique.